\documentclass{IEEEtran}
\usepackage{amsmath,amsfonts}
\usepackage{algorithmic}
\usepackage{algorithm}
\usepackage{array}
\usepackage[caption=false,font=normalsize,labelfont=sf,textfont=sf]{subfig}
\usepackage{textcomp}
\usepackage{diagbox}
\usepackage{stfloats}
\usepackage{url}
\usepackage{amsmath}
\usepackage{diagbox}
\usepackage{color}
\usepackage{makecell}
\usepackage{verbatim}
\usepackage{graphicx}
\usepackage{multirow}
\usepackage{cite}
\usepackage{epstopdf}
\usepackage{dcolumn}
\usepackage{booktabs}

\def\BibTeX{{\rm B\kern-.05em{\sc i\kern-.025em b}\kern-.08em
T\kern-.1667em\lower.7ex\hbox{E}\kern-.125emX}}
\begin{document}

\title{Cross-IQA: Unsupervised Learning for Image Quality Assessment}
\author{Zhen Zhang, Xiaolong Jiang, Chenyi Zhao, Shuang Qiao, and Tian Zhang

\thanks{Manuscript received XXXX, XX, 2024. This work was supported in part by the National Natural Science Foundation of China under Grants 11905028, 12105040 and in part by the Scientific Research Project of Education Department of Jilin Province under Grant JJKH20231294KJ. (Corresponding author: Tian Zhang).}

\thanks{Zhen Zhang, Xiaolong Jiang, Chenyi Zhao, Shuang Qiao and Tian Zhang are with the School of
Physics, Northeast Normal University, Changchun 130024, China (e-mail:zhangt100@nenu.edu.cn).}
}

\maketitle

\begin{abstract}
Automatic perception of image quality is a challenging problem that impacts billions of Internet and social media users daily. To advance research in this field, we propose a no-reference image quality assessment (NR-IQA) method termed Cross-IQA based on vision transformer(ViT) model. The proposed Cross-IQA method can learn image quality features from unlabeled image data. We construct the pretext task of synthesized image reconstruction to unsupervised extract the image quality information based ViT block. The pretrained encoder of Cross-IQA is used to fine-tune a linear regression model for score prediction. Experimental results show that Cross-IQA can achieve state-of-the-art performance in assessing the low-frequency degradation information (e.g., color change, blurring, etc.) of images compared with the classical full-reference IQA and NR-IQA under the same datasets.
\end{abstract}


\section{Introduction}

With the advent of the mobile Internet era, a great number of digital images are shared daily on social media platforms such as Instagram, Snapchat, Flickr, and others. Image quality score, an essential metric of digital images~\cite{ref1}, can help operators screen and provide high-quality digital images to their subscribers. Therefore, developing an image quality assessment (IQA) method that is highly consistent with the subjective perception of human vision becomes extremely important~\cite{ref2}.

Since the IQA technique can predict the quality scores of digital images, it is also widely used in a series of digital image processing tasks such as image restoration~\cite{ref3}, image super-resolution reconstruction~\cite{ref4}, and so on. Based on dependencies with reference images, IQA can be categorized into full-reference IQA (FR-IQA), reduced-reference IQA (RR-IQA), and no-reference IQA (NR-IQA). So far, FR-IQA is the most widely used IQA method (e.g., peak signal to noise ratio (PSNR), structural similarity index measure (SSIM), etc.)~\cite{ref5}, which can obtain the quality score by calculating the similarities and differences of the distorted image and the reference image. Unlike FR-IQA requiring the complete reference image, RR-IQA evaluates the quality of the distorted image using only partial information of the reference image (e.g., image entropy, gray histogram, and some transform-domain parameters), which can enhance the flexibility of IQA methods~\cite{ref6}. However, ideal reference images are usually not available in many real photographic environments. Therefore, the more challenging NR-IQA becomes a research hotpot\cite{ref7}.

Initially, traditional NR-IQA methods are mainly based on the assumption that the original image has a specific statistical distribution and the distortions alter this underlying distribution~\cite{ref8}. Therefore, feature extraction methods designed for some special features are used to evaluate the quality of images. Limited by the artificial features and feature extraction methods, traditional NR-IQA methods still have room for improvement. As deep learning (DL) technology has obtained remarkable advancements in various visual tasks such as image classification~\cite{ref9}, semantic segmentation~\cite{ref10}, and image recognition~\cite{ref11}, researchers have started to explore the DL-based NR-IQA methods~\cite{ref12,ref13,ref14}. Making full use of the excellent capability of feature extraction, DL-based NR-IQA achieves the significantly improved accuracy compared with those traditional manual feature extraction-based NR-IQA methods. Although the DL-based NR-IQA method has achieved remarkable success, the mainstream NR-IQA methods are based on supervised learning, which require large-scale image datasets with quality labels. Actually, providing quality scores to those datasets is indeed labor-intensive and costly. Therefore, the lack of large-scale IQA datasets is still a problem for the application of supervised learning-based NR-IQA.

To solve these problems, we propose a vision transformer (ViT)-based unsupervised image quality assessment method in this letter. Although the annotation process for the large-scale dataset is difficult, digital images with different degradation levels can be easily synthesized. Based on this, we design a pretext task of reconstructing images with varying levels of quality by training two parameter-shared encoders and two parameter-shared decoders. The well-trained encoder can effectively extract the image quality information in an unsupervised environment.

\section{Method}

\begin{figure*}[htbp]
	\centering
	\includegraphics[width=16cm]{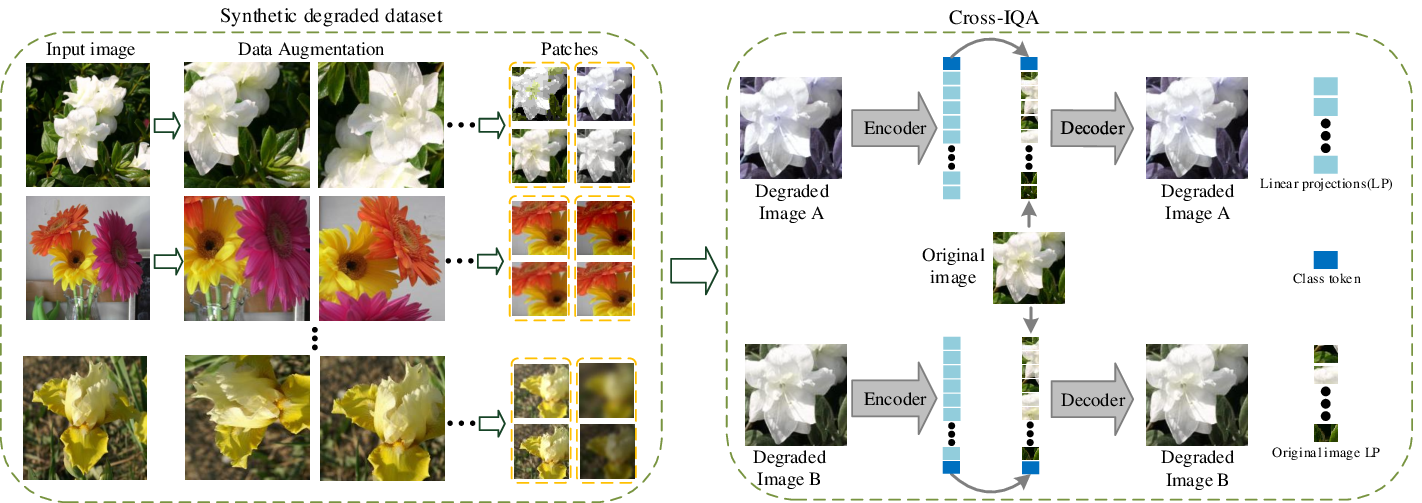}
	\caption{Schematic diagram of the proposed Cross-IQA.}
	\label{fig1}
\end{figure*}

\subsection{Cross-IQA}

Cross-IQA is a novel method designed for extracting image quality information in an unsupervised environment by image reconstruction and information exchange. Unlike those supervised learning-based NR-IQA using the Imagenet database, the proposed Cross-IQA model is deployed on a synthetic degraded dataset based on the Waterloo database. The schematic diagram of the proposed Cross-IQA is shown in Fig. \ref{fig1}. Specifically, the first two degraded images with different quality levels are put through two parameter-shared encoders for feature extraction. The obtained class token and linear projection of the original image are thereafter fed into the decoder for reconstructing the input degraded images. In the end, the well trained encoder can be used as a backbone for extracting the quality information to realize the subsequent no-reference image quality assessment.

\subsubsection{Cross-IQA encoder}

The standard ViT block~\cite{ref15} is used to construct the encoder of Cross-IQA. Firstly, positional embeddings and class token are first added to the input linear projections of the synthetic degraded images. Then, the processed linear projections are fed into a series of Transformer blocks for feature extraction. The reason why we choose ViT as the encoder and decoder can be summarized as:(i)ViT can efficiently extract features to class token, which is more convenient for feature exchange; (ii)The ViT-based encoder and decoder do not require skip connections, which is more convenient for the subsequent linear regression model.

\subsubsection{Cross class token}

After obtaining the class token by the encoder, it is thereafter connected to the linear projections of the original image to reconstruct the degraded images. This operation realizes the exchange of image quality information during the process of image reconstruction. It is an essential step for the proposed Cross-IQA method to unsupervised extract the expected image quality information.

\subsubsection{Cross-IQA decoder}

The ViT block used for decoder is the same as the encoder but the number is different. The constructed decoder is used to realize the image reconstruction task. The input of the decoder consists of the linear projection of the original image and exchanged class token. Note that, the decoder only performs the image reconstruction task in the pre-training phase of Cross-IQA.

\subsubsection{Reconstruction target}

For the reconstruction task, we use a loss function of mean square error (MSE) to quantify the discrepancy between the reconstructed and original images.

\begin{equation}
\label{MSE}
MSE = \frac{{\sum\limits_{i = 1}^n {{{\left( {y - f({x_i})} \right)}^2}} }}{n},
\end{equation}
where MSE represents the average of the squared distances between the predicted value of $f(x)$ and the ground truth of $y$.

At the same time, to better realize the exchange and extraction of image quality information in the proposed Cross-IQA framework, the mean absolute error (MAE) is also employed as a metric to calculate the difference between the input two degraded images and two reconstructed images.

\begin{equation}
\label{MAE}
MAE = \frac{{\sum\limits_{i = 1}^n {{\left( {y - f({x_i})} \right)}} }}{n},
\end{equation}
where MAE denotes the average distance between the predicted value of $f(x)$ and the the ground truth of $y$.

\begin{table*}
\begin{center}
\caption{PLCC comparison on individual distortion types.}
\label{tab1}
\begin{tabular}{c  l  c c c c c c c c c}

\hline
   & & \multicolumn{2}{c}{FR-IQA} & \multicolumn{7}{c}{NR-IQA}  \\
\hline
Database& \multicolumn{1}{c}{Type} & PSNR & SSIM & BRISQUE & CORNIA & GMLOG & DIQA & RANK & DFIA & \textbf{Cross-IQA}\\

\hline

TID2013 &$\# 01$ AGN& 0.934 &0.867 &0.630 &0.550 &0.748 &\textcolor{red}{0.915} &0.667 &0.718 & 0.641\\
        &$\# 02$ ANC &0.867 &0.773 &0.424 &0.209 &0.591 &\textcolor{red}{0.755} &0.620 &0.728 & 0.597\\
       & $\# 03$ SCN &0.916 &0.852 &0.727 &0.717 &0.769 &\textcolor{red}{0.878}& 0.821 &0.839& 0.712\\
        &$\# 04$ MN &0.836 &0.777 &0.321 &0.360 &0.491 &\textcolor{red}{0.734} &0.365 &0.581 & 0.597\\
        &$\# 05$ HFN &0.913 &0.863 &0.775& 0.797 &0.875 &\textcolor{red}{0.939} &0.760 &0.830 & 0.575\\
        &$\# 06$ IN &0.900 &0.750 &0.669 &0.585 &0.693 &0.843 &0.736 &\textcolor{red}{0.914} & 0.653\\
       & $\# 07$ QN &0.875 &0.866 &0.592 &0.727 &0.833 &0.858 &0.783 &0.824 & \textcolor{red}{0.861}\\
        &$\# 08$ GB &0.910 &0.967 &0.845 &0.840 &0.878 & 0.920 &0.809 &0.854 &\textcolor{red}{0.995}\\
       & $\# 09$ DEN &0.953 &0.925 &0.553 &0.721 &0.721 &0.788 &0.767 &0.821 & \textcolor{red}{0.972}\\
        &$\# 10$ JPEG &0.922 &0.920 &0.742 &0.806 &0.823 &0.892 &0.866 &\textcolor{red}{0.902} & 0.632 \\
       & $\# 11$ JP2K &0.886 &0.947 &0.799 &0.800 &0.872 &0.912 &0.878 &0.924 & \textcolor{red}{0.957}\\
        &$\# 12$ JGTE &0.806 &0.845 &0.301 &0.595 &0.400 &0.861 &0.704 &0.767 & \textcolor{red}{0.914}\\
       & $\# 13$ J2TE &0.891 &0.883 &0.672 &0.654 &0.731 &0.812 &0.810 &\textcolor{red}{0.859} & 0.754\\
       & $\# 14$ NEPN &0.679 &0.782 &0.175 &0.157 &0.190 &\textcolor{red}{0.659} &0.512 &0.455 & 0.31\\
       & $\# 15$ Block &0.330 &0.572 &0.184  &0.016 &0.318 &0.407 &0.622 &\textcolor{red}{0.683} & 0.574\\
       & $\# 16$ MS &0.757 &0.775 &0.155 &0.177 &0.119 &0.299 &0.268 &0.331 & \textcolor{red}{0.712} \\
       & $\# 17$ CTC &0.447 &0.378 &0.125 &0.262 &0.224 &0.687 &0.613 &0.761 & \textcolor{red}{0.891}\\
       & $\# 18$ CCS &0.634 &0.414 &0.032 &0.170 &-0.121 &-0.151 &0.662 &0.685 & \textcolor{red}{0.777}\\
       & $\# 19$ MGN &0.883 &0.780 &0.560 &0.407& 0.701 &\textcolor{red}{0.904}& 0.619 &0.700& 0.52\\
       & $\# 20$ CN &0.841 &0.857 &0.282& 0.541 &0.202 &0.655 &0.644 &0.725 & \textcolor{red}{0.747}\\
       & $\# 21$ LCNI &0.916 &0.806 &0.680 &0.696 &0.664 &0.930 &0.800 &0.829& \textcolor{red}{0.945}\\
       & $\# 22$ ICQD &0.820 &0.854 &0.804 &0.649 &0.886 &\textcolor{red}{0.936} &0.779 &0.837 & 0.652\\
       & $\# 23$ CHA &0.880 &0.878 &0.715 &0.689 &0.648 &0.756 &0.629 &0.780 & \textcolor{red}{0.962}\\
       & $\# 24$ SSR &0.911 &0.946 &0.800 &0.874 &0.915 &0.909& 0.859 &0.915 & \textcolor{red}{0.968}\\
       &   Low Frequency &0.831	&0.815	&0.521&	0.611	&0.565	&0.723&	0.721	&0.784	&\textcolor{red}{0.863}\\
       &   Other &0.808	&0.787	&0.527	&0.445	&0.626	&\textcolor{red}{0.797}	&0.65	&0.729	&0.583
\\
\hline
\hline
\end{tabular}
\end{center}
\end{table*}

\begin{figure}[htbp]
	\centering
	\includegraphics[width=8.5cm]{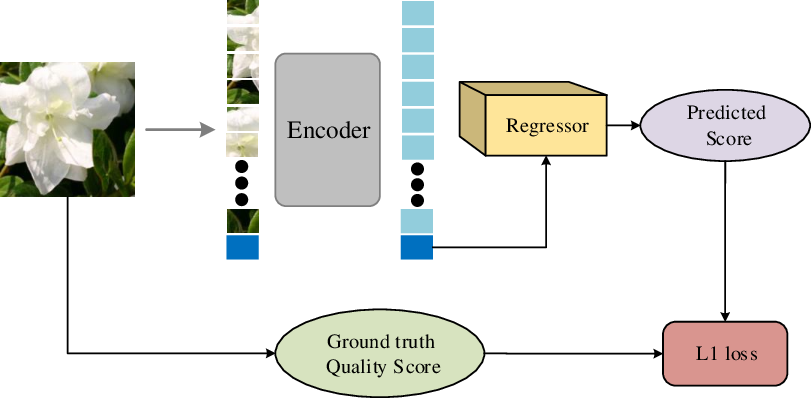}
	\caption{Process of Cross-IQA Regression.}
	\label{fig2}
\end{figure}

\begin{figure}[htbp]
	\centering
	\includegraphics[width=8.5cm]{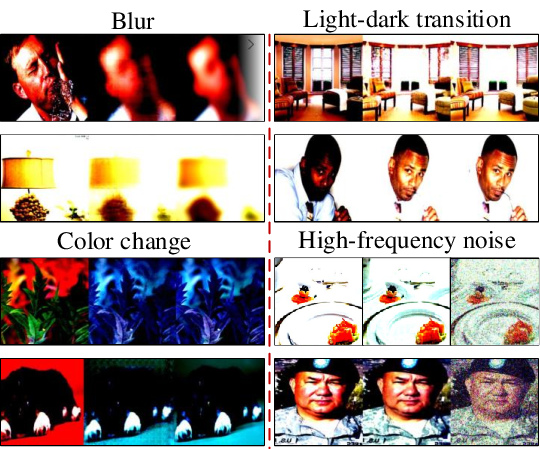}
	\caption{
      Example results on Waterloo database validation images. For each triplet, we show the original image (left), the reconstructed image by decoder of Cross-IQA (middle), and the synthetic degraded image (right). }
	\label{fig3}
\end{figure}

\subsection{Cross-IQA Regression}

After the pre-training phase, the well trained encoder of Cross-IQA is frozen. Then, the standard IQA database is thereafter used to train a regressor to map the obtained features by the encoder to the final perceptual image quality scores and the loss function is set as L1 loss, as shown in Fig. \ref{fig2}. Note that, the obtained class token by the encoder is the regressor head. Two public databases, including LIVE and TID2013 are used to fine-tune the linear regression model for score prediction~\cite{ref16,ref17}. LIVE contains 779 distorted images synthesized from 29 reference images while TID2013 comprises 3000 distorted images synthesized from 25 reference images with 24 synthetic distortion types and 5 degradation levels for each reference image.

\section{EXPERIMENTS}

The experiments are carried out on a workstation with a 8-core AMD Ryzen 7 CPU and an NVIDIA GeForce RTX 4090 GPU. We compared the performance of the proposed Cross-IQA method with the state-of-the-art IQA methods on public databases.

\subsection{Implementation details}

\textbf{Pre-training.} The proposed Cross-IQA was pre-trained on the synthetic degraded dataset (i.e., Waterloo database) described below. After balancing the time and resource constraints, we train the encoder of Cross-IQA by 200 epochs. The input image is randomly cropped to a size of $224 \times 224$ pixels and undergoes a random folding with the probability of 0.5 and normalized operation. AdamW is used as the optimizer, which is the common choice for ViT training. The number of ViT blocks is set to 12 and 8 respectively for the encoder and decoder. The batchsize is set to 16 and the learning rate is ${\rm{0}}{\rm{.0001*batchsize/256}}$. Fig. \ref{fig3} illustrates the original image, the reconstructed image by the decoder of Cross-IQA, and the synthetic degraded image, which indicates the proposed Cross-IQA is more sensitive to the low-frequency degradation information (e.g., color change, blurring, light-dark transition, etc.) than the high-frequency information.

\textbf{Fine-tuning.} In our experiment, the distorted images of a target IQA database (e.g., LIVE, TID2013) are divided into two portions: 80{\rm{\%}} is used for finetuning the Cross-IQA, and the remainder is used for testing. Note that, the database is divided according to the reference image to guarantee image content independence between the training and testing sets. Besides, the experiments are repeated ten times to obtain a fair evaluation, and the mean values of Spearman rank-order correlation coefficient (SROCC) and Pearsons linear correlation coefficient (PLCC) are reported as the final results~\cite{ref5}. All training samples are randomly flipped left and right with a probability of 0.5 and cropped to images with the size of $224 \times 224$ for data augment.

\begin{table}
\begin{center}
\caption{Generalization ability of GB pretrained-based Cross-IQA on TID2013.}
\label{tab2}
\begin{tabular}{c   c c c}
\hline

Pre-training& \multicolumn{1}{c}{Distortions with blur property} &  PLCC&  SROCC\\
\hline
           \multirow{9}*{Training in GB}
                & $\# 07$ QN&  0.861&0.854\\
                & $\# 08$ GB&  0.995&0.988\\
                & $\# 09$ DEN&  0.963&0.916\\
                & $\# 11$ JP2K&  0.911 & 0.871\\
                 & $\# 12$ JGTE&  0.914 & 0.785\\
                 & $\# 20$ CN &0.747 &0.750\\
                 & $\# 21$ LCNI &0.945 &0.891\\
                 & $\# 23$  CHA &0.967&0.921\\
                 & $\# 24$ SSR &0.968&0.905\\

\hline

\end{tabular}
\end{center}
\end{table}

\begin{table}
\begin{center}
\caption{PLCC and SROCC comprarion on the LIVE database.}
\label{tab3}
\begin{tabular}{c|   c c c c c c}
\hline

& PLCC &JP2K &JPEG &GN &GB &FF\\
\hline
           \multirow{3}*{\rotatebox{90}{FR-IQA}} &PSNR &0.873 &0.876 &0.926 &0.779 &0.87\\
                &SSIM~\cite{ref18} & 0.921 &0.955 &0.982& 0.893 &0.939\\
                & FSIM~\cite{ref19} & 0.91 &0.985 &0.976 &0.978 &0.912\\
           \hline

           \multirow{9}*{\rotatebox{90}{NR-IQA}} &DIVINE~\cite{ref20} &0.922& 0.921 &0.988 &0.923 &0.888\\

               &BLIINDS-II~\cite{ref21} & 0.935 &  0.968  & 0.98 &  0.938 &  0.896 \\
                &BRISQUE~\cite{ref22} &0.923 &0.973 &0.985 &0.951 &0.903\\
                &CORNIA~\cite{ref23} & 0.951 &0.965& 0.987 &0.968 &0.917\\
                 & CNN~\cite{ref24} & 0.953 &\textcolor{red}{0.981} &0.984 &0.953 &0.933\\
                 &SOM~\cite{ref25} &0.952 &0.961 &\textcolor{red}{0.991} &0.974 &0.944\\

                 \cline{2-7}
                 &\textbf{Cross-IQA} &\textcolor{red}{0.973} &0.792 &0.812 &\textcolor{red}{0.991} &\textcolor{red}{0.965}\\
\hline
\hline

& SROCC &JP2K &JPEG &GN &GB &FF\\
\hline
           \multirow{3}*{\rotatebox{90}{FR-IQA}} &PSNR & 0.87& 0.885 &0.942& 0.763 &0.874\\
                &SSIM~\cite{ref18} & 0.939 &0.946 &0.964& 0.907 &0.941\\
                & FSIM~\cite{ref19} & 0.97 &0.981 &0.967 &0.972 &0.949\\
           \hline

           \multirow{9}*{\rotatebox{90}{NR-IQA}} &DIVINE~\cite{ref20} & 0.913 &0.91 &0.984 &0.921 &0.863\\

               &BLIINDS-II~\cite{ref21} &  0.929 &0.942 &0.969 &0.923 &0.889 \\
                &BRISQUE~\cite{ref22} &0.914 &0.965 &0.979 &0.951 &0.887\\
                &CORNIA~\cite{ref23} & 0.943 & 0.955 & 0.976 & 0.969 & 0.906\\
                 & CNN~\cite{ref24} & 0.952 &\textcolor{red}{0.977} &0.978 &0.962 &0.908\\
                 &SOM~\cite{ref25} & 0.947 &0.952 &\textcolor{red}{0.984} &0.976 &0.937\\

                 \cline{2-7}
                 &\textbf{Cross-IQA} &\textcolor{red}{0.971} &0.755 &0.769 &\textcolor{red}{0.985} &\textcolor{red}{0.941}\\
\hline

\end{tabular}
\end{center}
\end{table}
\subsection{Performance comparison}

\textbf{Evaluation on TID2013.} For testing on TID2013, we generated 17 out of a total of 24 distortions (except for $\# 03$, $\# 04$, $\# 07$, $\# 12$, $\# 13$, $\# 20$, $\# 21$, and $\# 24$ as shown in Table \ref{tab1}). For the distortions that we could not generate, we fine-tuned the network according to the other trained distortions. During the testing process for each distortion type, only the corresponding synthetic distortion or similar distortion datasets with 5 distortion levels are used for the Cross-IQA pre-training. The original images used for constructing the synthetic degraded dataset were derived from the Waterloo database. For the TID2013 dataset with low-frequency distortion types (i.e.,$\# 07$ $\# 08$ $\# 09$ $\# 10$ $\# 11$ $\# 12$ $\# 13$ $\# 16$ $\# 17$ $\# 18$ $\# 20$ $\# 21$ $\# 23$ $\# 24$), our Cross-IQA method substantially outperforms the current state-of-the-art NR-IQA method in terms of PLCC, and matches or even exceeds the results of some FR-IQAs for some distortion types. Next, the Cross-IQA is pretrained with only the synthetic distortion of $\#08$ Gaussian blur (GB), the PLCC and SROCC results on the distortion types containing blur property from TID2013 dataset further demonstrate its generalization ability as shown in Table \ref{tab2}.

\textbf{Evaluation on LIVE.} To test on the LIVE database, we generated the training datasets with five distortion types including GB, Gaussian noise (GN), JPEG, JP2K and fast-fading (FF). The pre-training method used for LIVE was the same as the above TID2013. Table 3 shows that Cross-IQA has a significant advantage over other methods in evaluating features with low-frequency degradation, such as GB, JP2K, FF~\cite{ref18,ref19,ref20,ref21,ref22,ref23,ref24,ref25}. That means the proposed Cross-IQA without any labeled data for pre-training outperforms the existing NR-IQA and state-of-the-art FR-IQA methods in low-frequency degradation assessment.

The main reason for such excellent results is that ViT and its variants are good at capturing the low-frequency information of visual data including the global shape and structure of a scene or object. On the contrary, high-frequency information such as local edges and textures are usually not skillful~\cite{ref15}. This can be explained intuitively: self-attention, the main operation used in ViTs for exchanging information between non-overlapping patch tokens is a global operation, which is more suitable to capture global information (i.e., low-frequency) than local information (i.e., high-frequency). Therefore, for the low-frequency degradation information, the proposed Cross-IQA outperforms the state-of-the-art NR-IQA algorithm.

\textbf{Baseline performance analysis.} In this section, different pre-training methods are chosen to validate the effectiveness of Cross-IQA including the ViT network initialized with random parameters termed ViT baseline, ViT
network pre-trained using ImageNet termed ViT ImageNet, and the proposed Cross-IQA as shown in Table \ref{tab4}. The fine-tuning methods used therein are all the same as in the experiment above. All experiments were performed ten times, and the average SROCC and PLCC were given in Table \ref{tab4}. The proposed unsupervised Cross-IQA achieves quite good results in the evaluation of low-frequency degradation features of images. Since the proposed pretraining method uses unlabeled data to realize IQA, it has good potential for NR-IQA method with limited IQA data.

\begin{table}
\begin{center}
\caption{SROCC and PLCC for ViT using different pre-training methods.}
\label{tab4}
\begin{tabular}{c  |  c c c c }
\hline
\multirow{2}*{Method}&  \multicolumn{2}{c}{TID2013}  &  \multicolumn{2}{c}{LIVE}\\
\cline{2-5}

   &    PLCC&  SROCC &    PLCC&  SROCC\\
\hline

              ViT(Baseline)  &  0.285 & 0.258 &  0.385 & 0.362\\
             ViT(ImageNet)   &  0.369 &0.385 &   0.587 &0.535\\
               ViT(Cross-IQA) &  0.863 &0.842&   0.976 & 0.965
\\

\hline

\end{tabular}
\end{center}
\end{table}

\section{Conclusions}

In this letter, we proposed a novel no-reference image quality assessment method termed Cross-IQA based on the vision transformer model, which can extract the image quality information from unlabeled image data. Extensive experimental results demonstrate that the proposed Cross-IQA outperforms the classical full-reference IQA and NR-IQA in terms of low-frequency degradation information of images, thus demonstrating its application potential.

\end{document}